%% file: root.tex
\documentclass[letterpaper, 10 pt, conference]{ieeeconf}  

\IEEEoverridecommandlockouts                              

\usepackage{url}

\usepackage{makeidx}         
\usepackage{graphicx}        
\usepackage{multicol}        
\usepackage{amsmath}
\usepackage{amssymb}
\usepackage{booktabs}
\usepackage{xspace}
\usepackage{color}
\usepackage[hang,flushmargin]{footmisc}
\usepackage{microtype}
\usepackage{float}
\usepackage{array}
\usepackage{multirow}
\usepackage{flushend}
\usepackage{cite}

\renewcommand{\baselinestretch}{0.98}

\input{preamble}
\def\ourmodel{\ourName\xspace}

\title{\LARGE \bf
    \ourmodel: Predictive Tactile Sensing for Robotic Manipulation using Efficient Low-Dimensional Signals.
}

\author{Abdallah Ayad, Adrian Röfer, Nick Heppert, Abhinav Valada
\thanks{Department of Computer Science, University of Freiburg, Germany.}%
\thanks{This work was funded by the Carl Zeiss Foundation with the ReScaLe project and the BrainLinks-BrainTools center of the University of Freiburg.}
}
    
\begin{document}

\maketitle

\begin{abstract}
   \input{content/0_abstract}
\end{abstract}

\input{content/1_introduction}
\input{content/3_method}

\input{content/4_data}
\input{content/5_experiments}
\input{content/6_conclusion}

\begin{footnotesize}
    \bibliographystyle{IEEEtran}
    \bibliography{sources.bib}
\end{footnotesize}
\end{document}

%% file: preamble.tex
\usepackage{cite}
\usepackage{amsmath,amssymb,amsfonts}
\usepackage{algorithmic}
\usepackage{graphicx}
\usepackage{textcomp}
\usepackage{xcolor}

\usepackage{booktabs}
\usepackage{outlines}
\usepackage{caption}
\usepackage{subcaption}
\usepackage[hang,flushmargin]{footmisc}
\usepackage{stmaryrd}

\usepackage{microtype}
\usepackage{hhline}
\usepackage{multirow}
\usepackage{siunitx}
\usepackage{makecell}
\usepackage{gensymb}
\usepackage{placeins}

\definecolor{cvprblue}{rgb}{0.21,0.49,0.74}
\usepackage[breaklinks,colorlinks]{hyperref}

\input{assets/math_cmds}

\input{assets/text_cmds}

\usepackage[capitalize]{cleveref}
\crefname{section}{Sec.}{Secs.}
\Crefname{section}{Section}{Sections}
\Crefname{table}{Table}{Tables}
\crefname{table}{Tab.}{Tabs.}
\crefname{algorithm}{Algo.}{Algos.}
\Crefname{algorithm}{Algorithm}{Algorithms}

\definecolor{red}{RGB}{255, 0, 0}   
\definecolor{orange}{RGB}{255, 77, 0}   
\definecolor{green}{RGB}{0, 128, 0}   
\definecolor{purple}{RGB}{160, 32, 240}   
\definecolor{lightblue}{RGB}{52, 155, 235}   
\definecolor{darkmagenta}{RGB}{204, 51, 139} 

\def\BibTeX{{\rm B\kern-.05em{\sc i\kern-.025em b}\kern-.08em
    T\kern-.1667em\lower.7ex\hbox{E}\kern-.125emX}}

\input{assets/variable_definitions}

\usepackage{pifont}

%% file: assets/math_cmds.tex
\usepackage{tensor}

\DeclareMathAlphabet{\mathcal}{OMS}{cmsy}{m}{n}

\newcommand{\mat}[1]{\ensuremath{\mathbf{#1}}}

\newcommand{\set}[1]{\ensuremath{\left\{#1\right\}}}

\newcommand{\tf}[3]{\tensor[^{#1}]{\mat{#2}}{_{#3}}}

%% file: assets/text_cmds.tex
\usepackage{xspace}

\newcommand{\etal}{\emph{et~al.}\xspace}
   
\renewcommand{\[}{\begin{equation}}
\renewcommand{\]}{\end{equation}}

%% file: assets/variable_definitions.tex
\newcommand{\ourName}{Imagine2touch}

\newcommand{\ourNameMath}{IT}

\newcommand{\realSpace}{\mathbb{R}}

\newcommand{\reskin}{ReSkin\xspace}

\newcommand{\allObjects}{O}

\newcommand{\allTouches}{N}

\newcommand{\touchLocation}{l}
\newcommand{\touchLocationPredicted}{\tilde{\touchLocation}}



%% file: content/0_abstract.tex
Humans seemingly incorporate potential touch signals in their perception. Our goal is to equip robots with a similar capability, which we term \ourmodel. \ourmodel aims to predict the expected touch signal based on a visual patch representing the area to be touched.
We use \reskin, an inexpensive and compact touch sensor to collect the required dataset through random touching of five basic geometric shapes, and one tool. We train \ourmodel on two out of those shapes and validate it on the ood. tool.
We demonstrate the efficacy of \ourmodel through its application to the downstream task of object recognition. In this task, we evaluate \ourmodel performance in two experiments, together comprising 5 out of training distribution objects. \ourmodel achieves an object recognition accuracy of $58\%$ after ten touches per object, surpassing a proprioception baseline.

%% file: content/1_introduction.tex
\section{Introduction}
\label{sec:intro}

Dexterous object manipulation requires the ability to identify an object and track its state throughout the task. Vision-based methods for object detection~\cite{lang2022robust}, pose estimation~\cite{chisari2023centergrasp}, and tracking~\cite{von2023treachery} have become ubiquitous in robotics. However, vision, standalone, is an unreliable modality: the view of the object can be partially or fully occluded during the motion.
Humans, on the other hand, are capable of manipulating objects under occlusion using their proprioceptive and tactile senses: we can find a pen in a backpack without looking, with the help of our ability to predict how possible objects feel. In this work, we seek to enable robots to perform similar visuo-tactile skills. For this purpose, we platform \reskin  \cite{bhirangi2021reskin}, a magnetic-based tactile sensor, that is low cost and compact compared to its vision based counter parts such as~\cite{yuan2017gelsight, lambeta2020digit,lepora2022digitac}. 

To enable such skills, our approach fuses tactile and vision modalities using a common embedding. This approach has been widely researched across multiple modalities for various tasks~\cite{younes2023catch, hurtado2022semantic,rustler_active_2022,smith_3d_2020,murali_touch_2023,falco_cross-modal_2017,falco2019transfer,lee_making_2020,zhong_touching_2023,mildenhall2021nerf,lee_touching_2019, yang_generating_2023,yang_binding_2024}.
Specifically, in this work, we contribute a cross-modal model for predicting tactile readings from corresponding depth-image patches and an object-recognition demonstrator in which we use our trained model in an ensemble of probabilistic models scheme. We show that despite the low dimensional sensor readings, our method is able to achieve competitive results on this task. We share code and model at \url{https://github.com/AbdallahAyman/Imagine2touch}.

\begin{figure}
        \centering
        \begin{subfigure}{\columnwidth}
            \centering
            \begin{subfigure}{0.3\columnwidth}
            \includegraphics[width=\columnwidth]{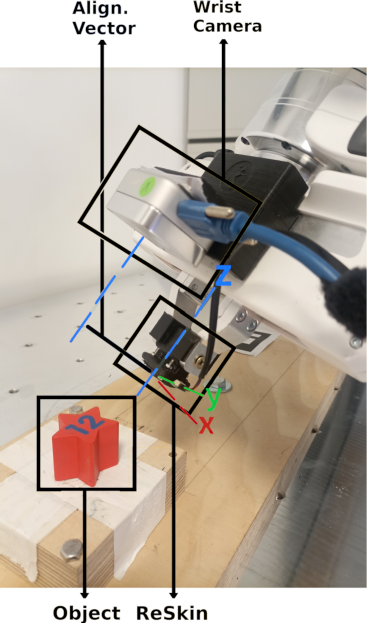}
            \caption{Robotic setup.}
            \label{fig:robotic_setup}
            \end{subfigure}
        \centering
        \begin{subfigure}{0.68\columnwidth}
            \centering
            \includegraphics[width=\columnwidth]{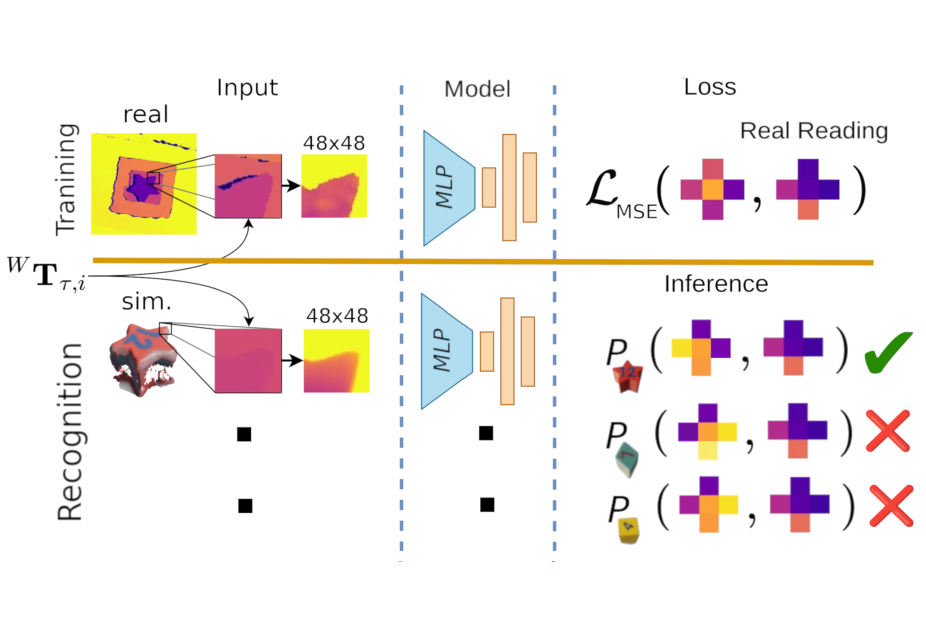}
            \caption{Data flow for training and object recognition.}
            \label{fig:setup_and_flow}
        \end{subfigure}
    \end{subfigure}
    \centering
    \begin{subfigure}{\columnwidth}
            \begin{subfigure}{0.3\columnwidth}
                \centering
                \includegraphics[width=\columnwidth]{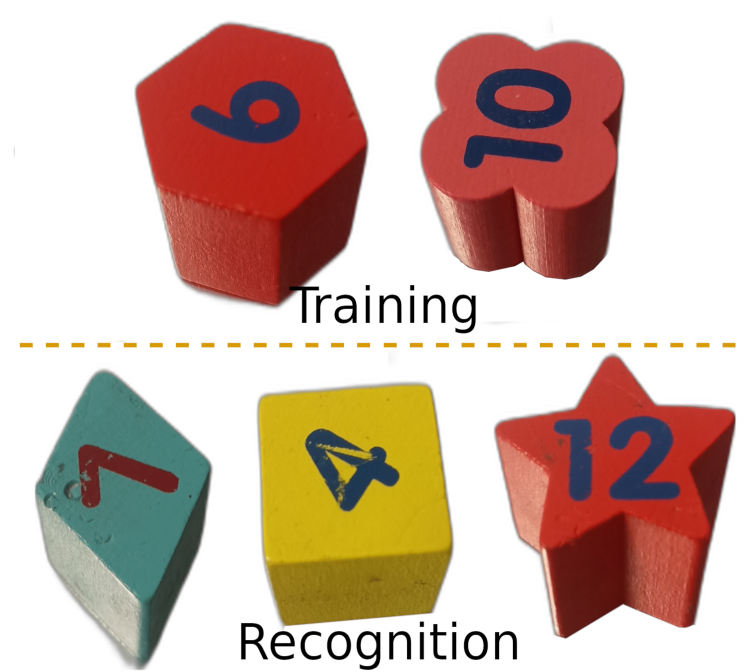}
                \caption{Objects: primitives.}
                \label{fig:pt_objects}
            \end{subfigure}    
            \begin{subfigure}{0.3\columnwidth}
                \centering
                \includegraphics[width=\columnwidth]{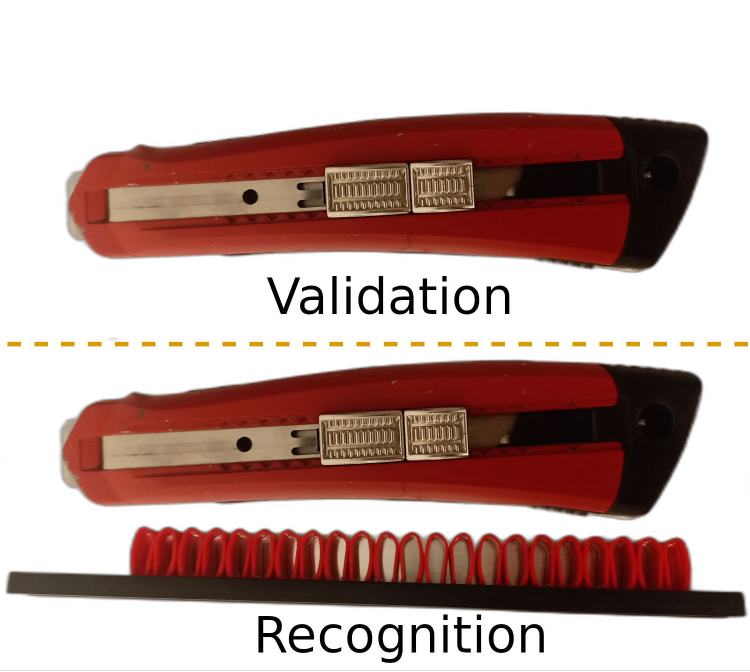}
                \caption{Objects: tools.}
                \label{fig:real_objects}
            \end{subfigure}
            \begin{subfigure}{0.3\columnwidth}
                \centering
                \includegraphics[width=\columnwidth]{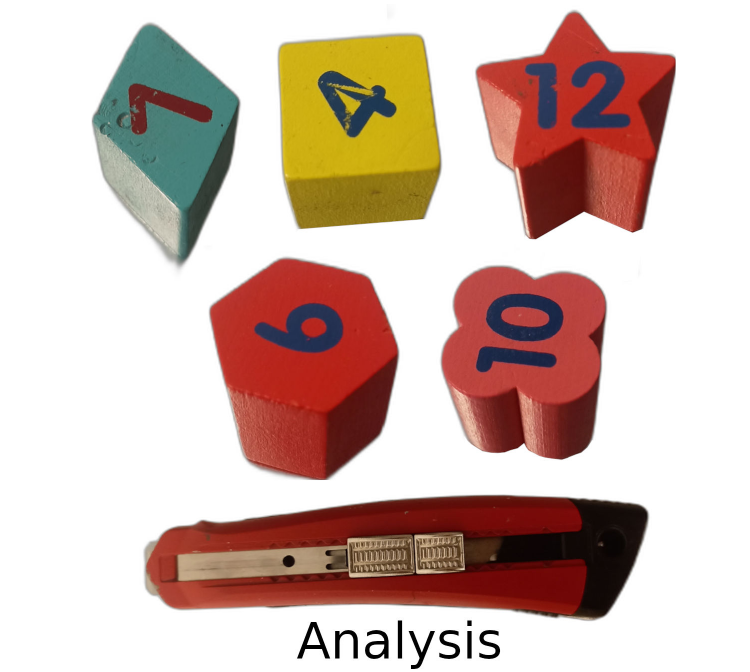}
                \caption{Full dataset.}
                \label{fig:objects_analysis}
            \end{subfigure}
    \end{subfigure}
    \centering
    \caption{
        \emph{(a)}: Robotic setup for our approach. The alignment vector shows the direction on which the robot moves for collecting one data sample to pair the wrist camera and ReSkin readings. \emph{(b)}: Data flow for training our model and using its inference in object recognition. The depth patch is cropped and processed from the full image using the end-effector pose $\tf{W}{T}{\tau,i}$ to match the touch area. It is then passed to the model, which we optimize using the MSE-loss between its output and the real touch reading. At recognition time, the robot has access only to possible 3D renderings. We use the probabilistic touch model in \cref{sec:technical_approach:object_recognition} for recognition. \emph{(c)}: Objects set: primitives. First row: primitives used for training the model. Second row: primitives used for one instance of the object recognition experiment. \emph{(d)}: Objects set: tools. First row: Tools used for validating the model. Second row: Tools used for the second instance of the object recognition experiment. \emph{(e)}: Full objects dataset used for analysis.
    }
\end{figure}

%% file: content/3_method.tex
\section{Technical Approach}
\label{sec:technical_approach}
We first detail our \ourName{} approach and then describe how we exploit it for the downstream task.
{\noindent \textbf{\ourName{}-Model}}:
\label{sec:technical_approach:our_model}
Our proposed model is a function $\ourNameMath : z_d \Rightarrow \tilde{\tau}$, which takes a processed depth image $z_d \in \realSpace^{48\times48}$ of the object surface as input, and predicts the tactile reading that would be emitted touching the surface $\tilde{\tau} \in \realSpace^{15}$.
We implement this function as a neural network, consisting of a single 200-neurons-layer MLP encoder, followed by a 5-neurons-bottleneck, the output of which is fed through a single 500-neurons-layer MLP decoder, see \cref{fig:setup_and_flow}. We add an auxiliary \emph{input-decoding} head with a separate 2000-neurons layer decoder to motivate modality fusion at the bottleneck stage.
We choose a low-capacity model to prevent overfitting on our small dataset and due to the size imbalance between the input and output of our network.
To train, validate, and analyze our model, we collect a dataset of tactile and visual pairs from 1630 samples from objects in \cref{fig:objects_analysis} using the setup in \cref{fig:robotic_setup}. The analysis shown in \cref{fig:data_tsne} indicates that mapping from one modality to the other should be possible.\looseness=-1

\label{sec:technical_approach:object_recognition}
{\noindent \textbf{Object Recognition}}: We use \ourmodel to recognize objects from a possible set. We subsequently perform $\allTouches$ touches, considering every touch as an independent probabilistic model. We update our object hypothesis based on the touch measurement through aggregating these models.

{\noindent \textit{Algorithmic Outline}}:
Let $\allObjects$ be a set of possible objects with known 3D representation (e.g. meshes/surfaces). Till we reach our maximum number of touches $\allTouches$, we calculate each object likelihood at step $i$ given the current observations consisting of previous readings $T = \set{\tau_1, \ldots, \tau_i}$. We sample from that distribution to get an object hypothesis $\tilde{o}$ on which we will sample a location $\touchLocationPredicted$  based on a heuristic that aims to distinguish $\tilde{o}$. We don't detail this heuristic here.

{\noindent \textit{Ensemble model}}: We assume that the probability of a possible object to be the true one after one touch follows a normal distribution with parameters ($\mu$) and ($\sigma^2$). We assume those parameters are included in the delta between the standardized actual touch signal ($\tau$) and the inferred touch signal ($\tilde{\tau}$).  We define our probabilistic model as follows:
\begin{equation}
P(o = o' | \tau_i, \tilde{\tau_i}) = e^{-(\tau_i - \tilde{\tau}_i)},
\end{equation}
where $o' \in \allObjects$ is a possible object and $o$ is the real object.
We assume independence between the touches to render the model more robust against noise and outliers. Additionally, we binarize the probabilities among possible objects by selecting a mutually exclusive winner to balance the weight of each model (i.e. touch). To finally recognize an object, akin to ensembling of weak models, the probability of an object can be calculated as the average among them:
\[ \small
P(o = o') = \frac{1}{N} (P(o' | \tau_i, \tilde{\tau_i}) + \ldots + P(o' | \tau_\allTouches, \tilde{\tau_\allTouches}))
\label{eq:object_probability}
\]

%% file: content/4_data.tex
\begin{figure}
    \centering
    \includegraphics[width=0.72\linewidth]{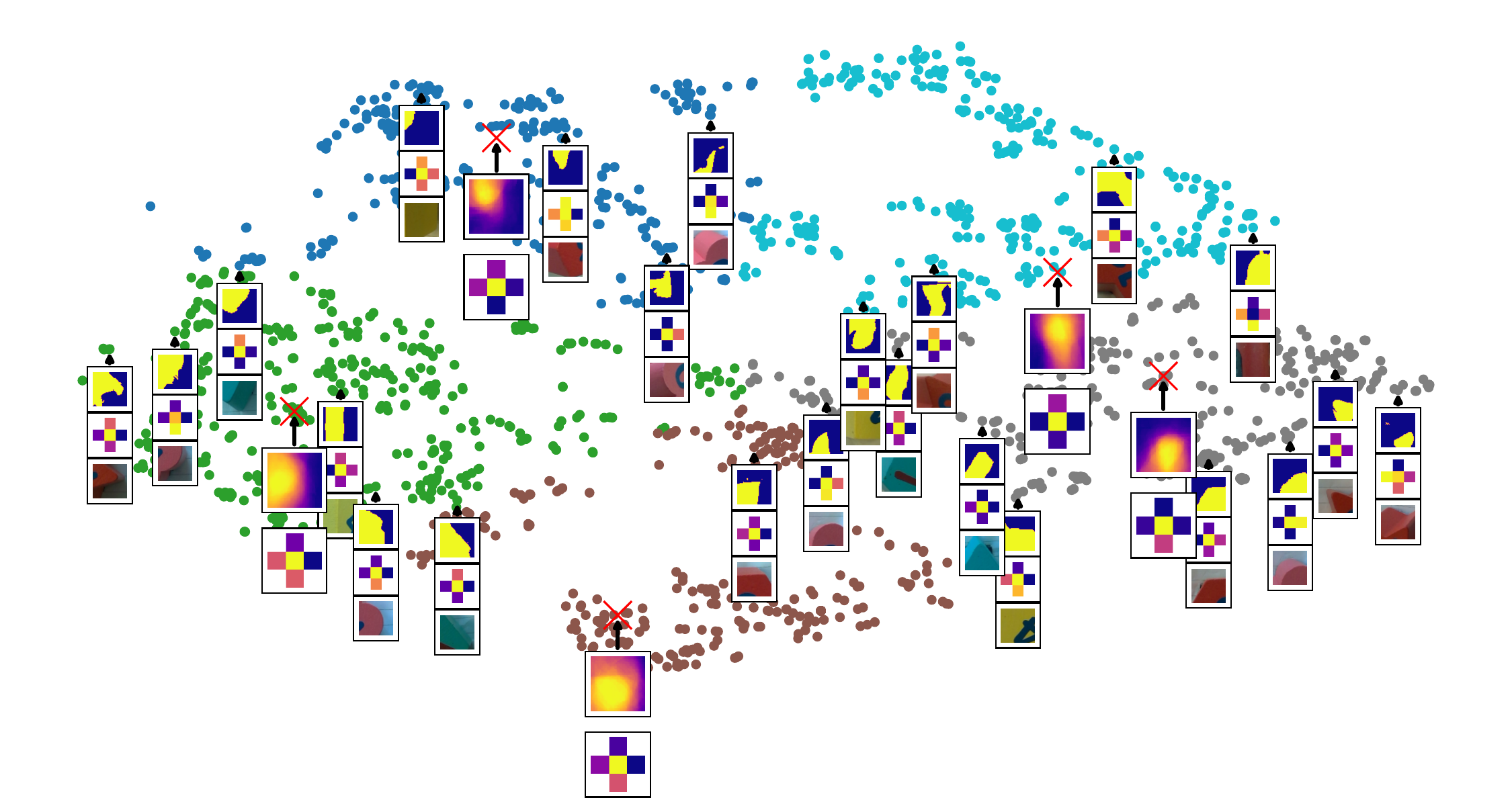}
    \caption{t-SNE plot of our data distribution. Five-means clustering of our processed depth data points with the associated images, tactile visualizations, and RGB images for example points.
    The means of the clusters are projected and highlighted in red with associated mean processed depth images and mean tactile visualizations.
    The plot shows distributed sensor activation, and correspondence between the depth patches and the signals.}
    \label{fig:data_tsne}
\end{figure}

%% file: content/5_experiments.tex
\section{Experiments}
\input{assets/figures/sanity_experiment}
\input{assets/tables/simple_recognition}

\input{assets/tables/recognition}
\label{sec:experiments}
We define two experiments. The first is shape classification. It is a feasibility check for the second. It demonstrates that higher concepts can be extracted from ReSkin sensor.

{\noindent \textbf{Shape Classification}}:
In the original work~\cite{bhirangi2021reskin}, Bhirangi~\etal demonstrates that it is feasible to identify exact touch locations and interaction forces.
To verify that the sensor can additionally be used for recognizing shapes (i.e. multi-touch contacts), we performed a preliminary experiment shown in \cref{fig:sanity_experiment}. We conclude that the sensor data can be used to distinguish between different contact shapes, see \cref{tab:sanity_evaluation}.

{\noindent \textbf{Object Recognition}}:
To evaluate \ourmodel performance against it, we define the following experiment. Akin to reaching into a backpack for a pen, in this task, the agent needs to identify the correct object out of a possible set, for which it has 3D models, on which \ourmodel{} predicts hypothetical touches. We use the ensembling scheme from \cref{sec:technical_approach:object_recognition} and compare \ourmodel performance to proprioception, which we define here as the delta bet. the real contact location and the nearest point in a possible 3D model. This baseline is the minimal tactile sense that could be implemented on any robot.

We conduct one instance for each out-of-training distribution primitive in Fig.~\ref{fig:pt_objects} and one for each tool in \ref{fig:real_objects}: The set $\allObjects$ in \cref{sec:technical_approach:object_recognition} is adapted for each instance. As we do not focus on pose estimation, we fix the objects to wooden bases to immobilize them.
The robot touches each object in each instance $10$ times in locations sampled according to our heuristic mentioned in \cref{sec:technical_approach:object_recognition}.
 We report the success rate per touch of recognizing the object in \cref{tab:recognition_results}. In conclusion, we find that \ourmodel improves over the proprioception performance in this task and generalizes to predicting touch signals outside of its training distribution.

%% file: assets/figures/sanity_experiment.tex
\begin{figure}
    \centering
    \includegraphics[width=1.2cm,angle=90]{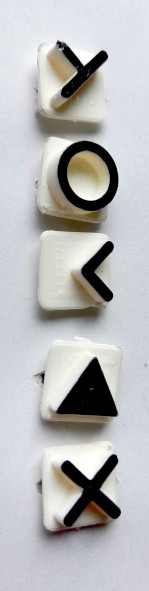}
    \caption{Shape classification experiment setup. We use differently shaped \emph{stamps} to indent the statically mounted sensor's gel pad in different locations. From left to right the shapes are: \emph{T}, \emph{circle}, \emph{angle}, \emph{triangle}, \emph{cross}. All stamps are at most $\SI{10}{\milli\meter}$ wide and $\SI{3.5}{\milli\meter}$ deep.}
    \label{fig:sanity_experiment}    
    \vspace{-0.2cm}
\end{figure}

%% file: assets/tables/simple_recognition.tex
\begin{table}[t]
    \centering
    \begin{tabular}{l|ccccc|c}
        \toprule
        Shape & Letter"T" & Circle & Angle & Triangle & Cross & Total  \\
        \midrule
        Acc. & 0.91 & 0.90 & 0.90 & 0.95 & 0.94 & 0.92 \\
        \bottomrule
    \end{tabular}
    \caption[Shape experiment results]{Shape experiment results. Shapes are shown respectively in \cref{fig:sanity_experiment}. Classification Accuracy Results for each shape and their average are produced using an MLP model with a hidden layer of size 500, a bottleneck of size 10, and a 5-class classification head. The model is trained on $80\%$ of $1280$ datapoints using cross-entropy loss.}
    \label{tab:sanity_evaluation}
    \vspace{-0.3cm}
\end{table}

%% file: assets/tables/recognition.tex
\begin{table}[t]
    \vspace{2mm}
    \setlength{\tabcolsep}{3pt} 
    \centering
    \begin{tabular}{l|ccccccccc}
        \toprule
        Object set && \multicolumn{2}{c}{Primitives}  && \multicolumn{2}{c}{Tools} && \multicolumn{2}{c}{\emph{Mean}}\\
        \midrule
        Touch Model
        &&  prop. & I2T  && prop. & I2T  && prop. & I2T\\
        \cmidrule{3-4} \cmidrule{6-7} \cmidrule{9-10}
           && $23\%$ & $\mathbf{50\%}$ && $\mathbf{80\%}$ & $70\%$ && $46\%$ & $\mathbf{58\%}$ \\
        \bottomrule
    \end{tabular}
\caption{Results from the object recognition experiment. We find that both I2T (Imagine2touch) and prop. (proprioception) work for identifying objects from the tools set. The primitive objects are more difficult for both methods due to their similarity in extent and tactile features. Despite the inherent difficulty of distinguishing similar objects without a dense measure such as vision, \ourName{} exceeds random chance, and the proprioception baseline across the objects sets.
}
\label{tab:recognition_results}
    \vspace{-0.2cm}
\end{table}

%% file: content/6_conclusion.tex
\section{Conclusion}
\label{sec:conclusion}
In this work, we investigated the use of the novel, low-cost, and compact \reskin sensor as a platform to learn a predictive touch sense for general robotic tasks. We proposed \ourmodel, a novel approach that infers expected tactile readings from small depth images of surfaces. We additionally introduce a procedure for collecting data to train the model. We leveraged the model for a downstream task involving five OOD objects and demonstrated that our model is able to generalize.
We view our results as an encouraging step towards using inexpensive tactile sensors such as \reskin more often in robotics.
For future work, we see an opportunity for building the inverse of our approach: a model predicting depth images from tactile signals to obtain 3D object features. This would enable full tactile 3D reconstruction.

%% file: root.bbl
\begin{thebibliography}{10}
\providecommand{\url}[1]{#1}
\csname url@samestyle\endcsname
\providecommand{\newblock}{\relax}
\providecommand{\bibinfo}[2]{#2}
\providecommand{\BIBentrySTDinterwordspacing}{\spaceskip=0pt\relax}
\providecommand{\BIBentryALTinterwordstretchfactor}{4}
\providecommand{\BIBentryALTinterwordspacing}{\spaceskip=\fontdimen2\font plus
\BIBentryALTinterwordstretchfactor\fontdimen3\font minus \fontdimen4\font\relax}
\providecommand{\BIBforeignlanguage}[2]{{%
\expandafter\ifx\csname l@#1\endcsname\relax
\typeout{** WARNING: IEEEtran.bst: No hyphenation pattern has been}%
\typeout{** loaded for the language `#1'. Using the pattern for}%
\typeout{** the default language instead.}%
\else
\language=\csname l@#1\endcsname
\fi
#2}}
\providecommand{\BIBdecl}{\relax}
\BIBdecl

\bibitem{lang2022robust}
C.~Lang, A.~Braun, and A.~Valada, ``Robust object detection using knowledge graph embeddings,'' in \emph{DAGM German Conference on Pattern Recognition}, 2022, pp. 445--461.

\bibitem{chisari2023centergrasp}
E.~Chisari, N.~Heppert, T.~Welschehold, W.~Burgard, and A.~Valada, ``Centergrasp: Object-aware implicit representation learning for simultaneous shape reconstruction and 6-dof grasp estimation,'' \emph{arXiv preprint arXiv:2312.08240}, 2023.

\bibitem{von2023treachery}
J.~O. von Hartz, E.~Chisari, T.~Welschehold, W.~Burgard, J.~Boedecker, and A.~Valada, ``The treachery of images: Bayesian scene keypoints for deep policy learning in robotic manipulation,'' \emph{IEEE Robotics and Automation Letters}, 2023.

\bibitem{bhirangi2021reskin}
R.~Bhirangi, T.~Hellebrekers, C.~Majidi, and A.~Gupta, ``Reskin: versatile, replaceable, lasting tactile skins,'' in \emph{Proc. Conf. on Rob. Learn.}, 2021.

\bibitem{yuan2017gelsight}
W.~Yuan, S.~Dong, and E.~H. Adelson, ``Gelsight: High-resolution robot tactile sensors for estimating geometry and force,'' \emph{{IEEE} Sensors}, vol.~17, no.~12, p. 2762, 2017.

\bibitem{lambeta2020digit}
M.~Lambeta, P.-W. Chou, S.~Tian, B.~Yang, B.~Maloon \emph{et~al.}, ``Digit: A novel design for a low-cost compact high-resolution tactile sensor with application to in-hand manipulation,'' \emph{IEEE Robotics and Automation Letters}, vol.~5, no.~3, pp. 3838--3845, 2020.

\bibitem{lepora2022digitac}
N.~F. Lepora, Y.~Lin, B.~Money-Coomes, and J.~Lloyd, ``Digitac: A digit-tactip hybrid tactile sensor for comparing low-cost high-resolution robot touch,'' \emph{IEEE Robotics and Automation Letters}, vol.~7, no.~4, pp. 9382--9388, 2022.

\bibitem{younes2023catch}
A.~Younes, D.~Honerkamp, T.~Welschehold, and A.~Valada, ``Catch me if you hear me: Audio-visual navigation in complex unmapped environments with moving sounds,'' \emph{IEEE Robotics and Automation Letters}, vol.~8, no.~2, pp. 928--935, 2023.

\bibitem{hurtado2022semantic}
J.~V. Hurtado and A.~Valada, ``Semantic scene segmentation for robotics,'' in \emph{Deep learning for robot perception and cognition}, 2022, pp. 279--311.

\bibitem{rustler_active_2022}
L.~Rustler, J.~Lundell, J.~K. Behrens, V.~Kyrki, and M.~Hoffmann, ``Active {Visuo}-{Haptic} {Object} {Shape} {Completion},'' \emph{IEEE Robotics and Automation Letters}, vol.~7, no.~2, pp. 5254--5261, 2022.

\bibitem{smith_3d_2020}
E.~Smith, R.~Calandra, A.~Romero, G.~Gkioxari, D.~Meger, J.~Malik, and M.~Drozdzal, ``{3D} {Shape} {Reconstruction} from {Vision} and {Touch},'' in \emph{Proc. Adv. Neural Inform. Process. Syst.}, 2020.

\bibitem{murali_touch_2023}
P.~K. Murali, B.~Porr, and M.~Kaboli, ``Touch if it's transparent! {ACTOR}: {Active} {Tactile}-based {Category}-{Level} {Transparent} {Object} {Reconstruction},'' in \emph{Proc. IEEE Int. Conf. on Intel. Rob. and Syst.}, 2023.

\bibitem{falco_cross-modal_2017}
P.~Falco, S.~Lu, A.~Cirillo, C.~Natale, S.~Pirozzi, and D.~Lee, ``Cross-modal visuo-tactile object recognition using robotic active exploration,'' in \emph{Proc. IEEE Int. Conf. on Rob. and Auto.}, 2017.

\bibitem{falco2019transfer}
P.~Falco, S.~Lu, C.~Natale, S.~Pirozzi, and D.~Lee, ``A transfer learning approach to cross-modal object recognition: from visual observation to robotic haptic exploration,'' \emph{IEEE Trans. Robot.}, 2019.

\bibitem{lee_making_2020}
M.~A. Lee, Y.~Zhu, P.~Zachares, M.~Tan, K.~Srinivasan, S.~Savarese, L.~Fei-Fei, A.~Garg, and J.~Bohg, ``Making {Sense} of {Vision} and {Touch}: {Learning} {Multimodal} {Representations} for {Contact}-{Rich} {Tasks},'' \emph{IEEE Trans. on Robotics}, vol.~36, no.~3, 2020.

\bibitem{zhong_touching_2023}
S.~Zhong, A.~Albini, O.~P. Jones, P.~Maiolino, and I.~Posner, ``Touching a {NeRF}: {Leveraging} {Neural} {Radiance} {Fields} for {Tactile} {Sensory} {Data} {Generation},'' in \emph{Proc. Conf. on Rob. Learn.}, 2023, pp. 1618--1628.

\bibitem{mildenhall2021nerf}
B.~Mildenhall, P.~P. Srinivasan, M.~Tancik, J.~T. Barron, R.~Ramamoorthi, and R.~Ng, ``Nerf: Representing scenes as neural radiance fields for view synthesis,'' \emph{Communications of the ACM}, vol.~65, no.~1, pp. 99--106, 2021.

\bibitem{lee_touching_2019}
J.-T. Lee, D.~Bollegala, and S.~Luo, ``"{Touching} to {See}" and "{Seeing} to {Feel}": {Robotic} {Cross}-modal {Sensory} {Data} {Generation} for {Visual}-{Tactile} {Perception},'' in \emph{Proc. IEEE Int. Conf. on Rob. and Auto.}, 2019.

\bibitem{yang_generating_2023}
F.~Yang, J.~Zhang, and A.~Owens, ``Generating visual scenes from touch,'' in \emph{Proc. IEEE Conf. Comput. Vis. Pattern Recog.}, 2023, pp. 22\,070--22\,080.

\bibitem{yang_binding_2024}
F.~Yang, C.~Feng, Z.~Chen, H.~Park, D.~Wang, Y.~Dou, Z.~Zeng, X.~Chen, R.~Gangopadhyay, A.~Owens, and A.~Wong, ``Binding {Touch} to {Everything}: {Learning} {Unified} {Multimodal} {Tactile} {Representations},'' \emph{arXiv preprint arXiv:2401.18084}, 2024.

\end{thebibliography}
